\pdfoutput=1
\documentclass[letterpaper, 10 pt, conference]{ieeeconf}

\IEEEoverridecommandlockouts                              

\usepackage{amsmath} 
\usepackage{amssymb}  
\usepackage{graphicx}
\usepackage[noend]{algpseudocode}
\usepackage{algorithmicx,algorithm}
\usepackage{color}

\title{\LARGE \bf
Navigation by Imitation in a Pedestrian-Rich Environment
}
\author{
  Jing Bi, Tianyou Xiao, Qiuyue Sun and Chenliang Xu
  \thanks{*This work was supported by the NSF Grants No. 1741472 and No. 1813709. Any opinions, findings, and conclusions or recommendations are those of the authors and do not necessarily reflect the views of the NSF.}
  \thanks{All authors are with the University of Rochester, Rochester, NY 14627, USA. Emails: jbi5@ur.rochester.edu, \{txiao3, qsun15\}@u.rochester.edu, chenliang.xu@rochester.edu.}
}

\begin{document}

\maketitle
\thispagestyle{empty}
\pagestyle{empty}

\begin{abstract}
Deep neural networks trained on demonstrations of human actions give robot the ability to perform self-driving on the road. However, navigation in a pedestrian-rich environment, such as a campus setup, is still challenging---one needs to take frequent interventions to the robot and take control over the robot from early steps leading to a mistake. An arduous burden is, hence, placed on the learning framework design and data acquisition. In this paper, we propose a new learning-from-intervention Dataset Aggregation (DAgger) algorithm to overcome the limitations brought by applying imitation learning to navigation in the pedestrian-rich environment. Our new learning algorithm implements an error backtrack function that is able to effectively learn from expert interventions. 
Combining our new learning algorithm with deep convolutional neural networks and a hierarchically-nested policy-selection mechanism, we show that our robot is able to map pixels direct to control commands and navigate successfully in real world without explicitly modeling the pedestrian behaviors or the world model. 
\end{abstract}

\section{INTRODUCTION}
\label{sec:intro}

Autonomous ground vehicle industry has gained great development from the recent successes in deep learning. Applications such as personal mobility services and luggage carrying support in complex, pedestrian-rich environment become feasible~\cite{bai2015intention, chen2017socially}. Although walking in crowded environment is not difficult for humans, it is actually a challenge to autonomous vehicles, since some subtle social norms that pedestrians usually follow are hard to quantify~\cite{chen2017socially}. Intuitively, employing imitation learning by mimicking human pedestrians is a promising approach to start with.

Imitation learning has been gaining increasing importance as a promising learning approach to achieve high performance on various applications, e.g., humanoid robots. Traditional imitation learning is based on behavior cloning, where robots learn from human acting trajectories in a supervised learning fashion. The human behavior implicitly provides the learner information about the states it might encounter and the possible actions it should take~\cite{goil}. Despite the simplicity, many researches have proved it to be an efficient and high-performance strategy~\cite{schaal1999imitation, abbeel2004apprenticeship, bagnell2007boosting, ross2010efficient}. 

However, learning solely from cloning states and actions induced by human behaviors might likely lead to failures when the robot encounters states that are never visited by human demonstrations. The Dataset Aggregation (DAgger) algorithm~\cite{ross2011reduction} is, therefore, introduced to alleviate this problem. The DAgger describes an improved learning paradigm, where states are sampled from the learner, after a warm-start with behavior cloning, and annotated along with expert behavior when encountering mistakes. These data are aggregated for training a better learner policy iteratively. The use of DAgger is, hence, preferred by many researchers for imitation learning. 

\begin{figure}[t]
  \centering
  \includegraphics[width=\linewidth]{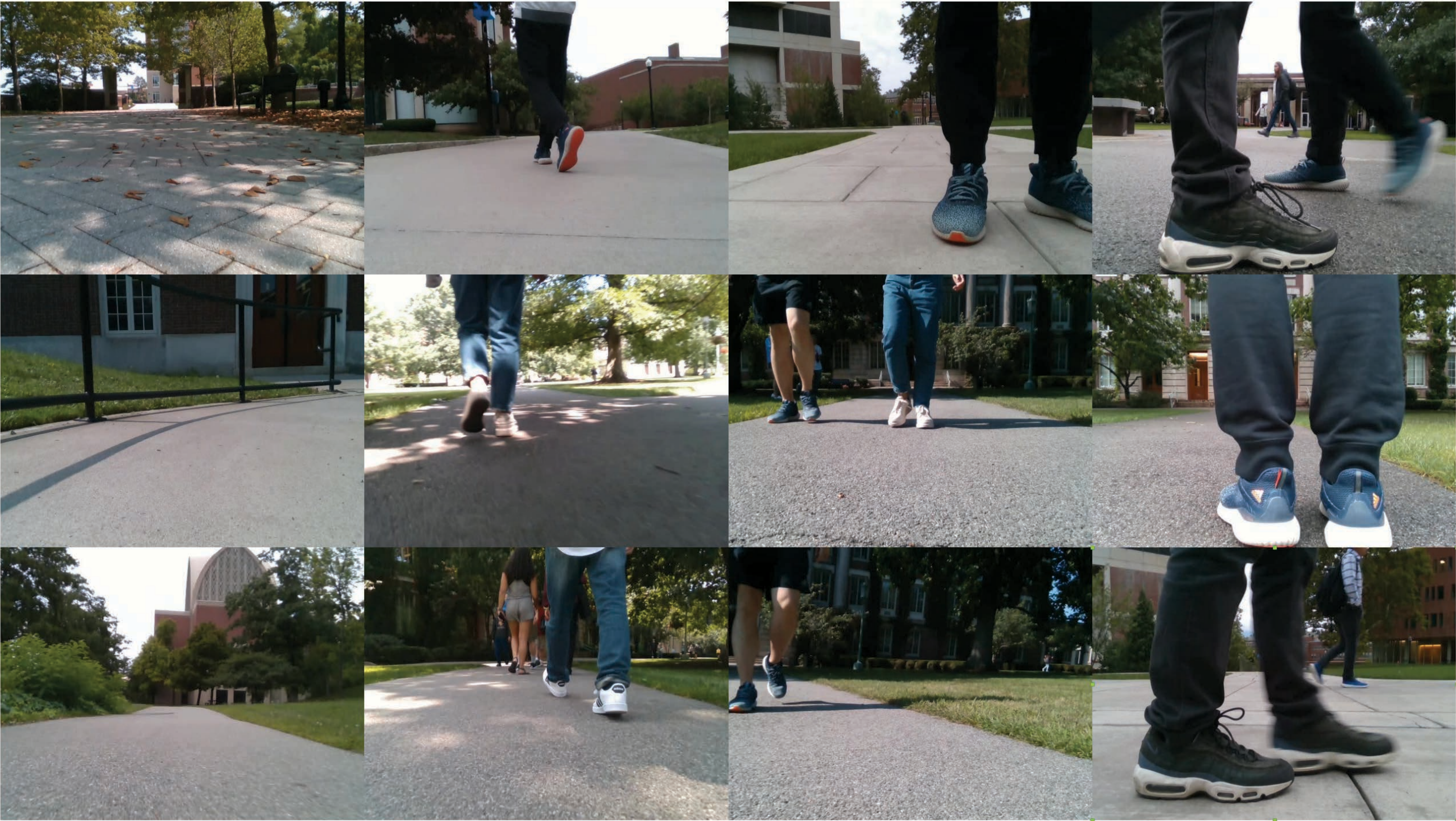}
  \caption{We train a robot to navigate in pedestrian-rich environment, where we consider four different scenarios (columns from left to right): \textit{no pedestrian (path following)}, \textit{confronting}, \textit{pedestrian following}, and \textit{crossing (avoid hitting)}.}
  \label{fig:dataset}
\end{figure}

In this paper, we consider to apply imitation learning to real-world robot navigation in a pedestrian-rich environment. Figure~\ref{fig:dataset} shows a snapshot of the navigation problems we encounter, where we roughly split all scenarios into four categories: \textit{no pedestrian (path following)}, \textit{confronting}, \textit{pedestrian following}, and \textit{crossing (avoid hitting)}. In this environment, we need to make frequent interventions to the robot and take control over the robot from early steps leading to a mistake. In other words, we cannot allow robot to execute actions generated by an imperfect policy, which might jeopardize the safety of pedestrians. Such requirements make it difficult to collect effective data trajectories using the DAgger algorithm for training. 


To address the above limitation, we propose a new Learn-from-Intervention DAgger algorithm that implements an error backtrack function. The algorithm can effectively learn from the pairs of data containing states sampled from the learner policy and the backtracked action losses. Compared with traditional DAgger implementation, our framework allows the human expert to interact with the robot, which largely raises the efficiency of Data Aggregation process. Combining our new learning algorithm with deep Convolutional Neural Networks (CNNs), such as ResNet~\cite{he2016deep}, we are able to map pixels direct to control commands  without explicitly modeling the pedestrian behaviors or the world model. We implement this new imitation learning framework on a novel mobile robot platform and experiments demonstrate promising performance in terms to navigation by imitation in a pedestrian-rich campus environment.

\section{RELATED WORK}
\label{sec:related}



A common approach to navigate in pedestrian-rich environment is to treat pedestrians as dynamic obstacles and use specific control strategy for robots to avoid collision \cite{fox1997dynamic, phillips2011sipp, van2011reciprocal}. Initially, researchers have to build sophisticated models for pedestrians and manually define actions to avoid accidents based on possible states the robot could enter. Unfortunately, it is extremely difficult to include all possible states in their implementation due to the complexity of real-world environment. When the robot encounters a situation that does not exist in the model, it will likely fail to find a feasible action, which leads to the \textit{freezing robot problem}. Several attempts~\cite{chen2017socially, socially} have been made to solve this problem, among which they focus on learning-based methods. Unlike previous hand-craft approaches, their implementation essentially gives the robot ability to learn and emulate human behaviors. A better policy could be generated by the robot itself instead of being hard-coded by developers. However, an explicit model for pedestrians is required in their training process (which is not required in our approach). In our model, human operators could easily evaluate whether actions of the robot is acceptable and time-efficient while driving it in the real-world environment. By interactively adjusting behaviors of the robot, human operators could expose their decision-making policies to it and from which the robot could derive its own policy based on imitation without explicitly modeling the behavior of pedestrians.

Imitation learning is one form of learning from demonstration~\cite{learn-from-demonstration} and has been applied to a variety of tasks, including autonomous flight~\cite{cil-1, cil-13, cil-30}, learning navigational behavior~\cite{cil-37, cil-38}, off-road driving~\cite{cil-22, cil-32}, and road following~\cite{cil-4, cil-6, cil-27, cil-36}. Main approaches for imitation learning can be categorized into behavior cloning (BC) and inverse reinforcement learning (IRL). Behavior cloning tackles this problem in a supervised manner, by directly learning the mapping from the states to corresponding actions labeled by the expert policy. This learning algorithm relies on the presence of an expert who instruct the agents at each state and the agents try to mimic the corresponding action. In~\cite{end-to-end}, Codevilla et al. use a CNN to directly map observation to steer angle, which give CNN ability to perform self-driving on road with different conditions. In~\cite{UAV}, imitation learning is applied to the navigation of a UAV in the forest without explicitly modeling the environment. Several studies~\cite{cil-27, cil-22, cil-4} have been conducted on the similar topic. Unlike our implementation, their approaches mainly focus on learning steering. We further expand the domain of tasks to include acceleration and braking along with a new learning-from-intervention DAgger algorithm, which empowers our robot to adapt to a more dynamic environment. Instead of concentrating on lane following and static obstacles avoidance, we develop a framework which could handle crowded environment including bypassing pedestrians.

\section{METHOD}
\label{sec:method}

In this paper, we study navigation in a pedestrian-rich environment and formulate it as an imitation learning problem. Instead of explicitly modeling the pedestrian behaviors as in~\cite{socially}, we split all scenarios into four categories: $\mathcal{G} =$ \{\textit{no pedestrian (path following)}, \textit{confronting}, \textit{pedestrian following}, \textit{crossing (avoid hitting)}\}. Each category has a task-specific policy $\{\pi_g: s \mapsto a | g \in \mathcal{G}\}$ that maps a state to a robot action; the latter is defined by speed and steering angle in a discrete space. A meta-controller $\mu: s \mapsto g$ decides to perform a particular task given the state. In our case, the states are observations obtained from cameras. To train a policy, we first initialize it via behavior cloning, which serves as \textit{warm-start}, then the policy is iteratively refined by our proposed \textit{Learning-from-Intervention DAgger} algorithm. Next, we introduce each technique in detail. 

\subsection{Hierarchical Formalism}

The meta-controller and the sub-task controllers are nested in a two-layer tree structure, where the meta-controller is the root and the four sub-task controllers serve as leaves. We learn the two levels of policies separately. We use CNNs to achieve our policy mappings from states to tasks/actions. Despite the separate policies, they share the same base network structure, i.e., ResNet in our implementation; hence, the added computation is minor. In testing, given the observations at each time step, the high-level policy $\mu$ decides on which tasks to select and the task policy $\pi_g$ generates an action for the robot.

\subsection{Hierarchical Behavior Cloning}

\begin{algorithm}[t]
\caption{Hierarchical Behavior Cloning} 
\hspace*{0.02in} {\bf Input:} parameters $T$\\
\hspace*{0.02in} {\bf Output:} initialized policies $\mu$, $\pi_g , \forall g \in \mathcal{G}$ 
\begin{algorithmic}[1]
\State Initialize data buffers $D_{h} \gets \emptyset$ and  $D_{g} \gets \emptyset$, $\forall g \in \mathcal{G}$ 
\State Initialize policies $\mu$, $\pi_g , \forall g \in \mathcal{G}$ 
\Repeat
\State Get a new environment instance with start state $s_0$
\For{$t=1,...,T$}
　　\State Get an updated state $s_t$
　　\State Append  $D_h \gets D_h \bigcup \{(s_t, \mu^*(s_t))\}$ 
　　\State Append  $D_g \gets D_g \bigcup \{(s_t, \pi^*(s_t))\}$ 
\EndFor
\Until End of collecting
\State Train meta-controller $\mu \gets$ Train $(\mu,D_h)$ 
\State Train sub-policies $\pi_g \gets$ Train $(\pi_g,D_{g})$ for all $g$
\end{algorithmic}
\label{alg:hbc}
\end{algorithm}

We begin by describing the standard imitation learning setup and then proceed to our DAgger algorithms. Consider a robot that interacts with the environment over discrete time steps. At each time step $t$, the robot receives an observation as its state representation $s_t$ and takes an action by $\pi_{\mu(s_t)}(s_t)$. The basic idea behind imitation learning is to learn controller polices that mimic an expert. Since the expert is successful at performing the task of interest, a robot trained to mimic expert actions will also perform the task well.

Let us define the expert knowledge of the meta-controller as $\mu^*$ and that of the sub-task controllers as $\pi^*$. Without explicitly model the behavior of the pedestrian, we drive our robot platform in real-world to collect the expert data. We use the hierarchical behavior cloning algorithm (see Algorithm~\ref{alg:hbc}) to initialize our controller policies. The algorithm collects separate state-task and state-action pairs from expert driving data to form datasets $D_h$ and $D_g$ to train meta-controller police and sub-task controller polices, respectively.

\subsection{Hierarchical Learn-from-Mistake DAgger} 

\subsubsection{DAgger} 
Using only behavior cloning, the best policy we can get is learned from the distribution of states generated by the expert, not the robot itself. Such policy might likely fail when the robot encounters a state that is never visited by the expert. The Dataset Aggregation (DAgger) algorithm is introduced in~\cite{ross2011reduction} to alleviate this problem. In its simplest form, DAgger works iteratively as follows. At the outset, we collect data $D^1 = \{(s, \pi^*(s)) | s \sim \pi^*\} $ and initialize policy $\pi^1$ by behavior cloning the expert demonstrations; here $s \sim \pi^*$ denotes the states are sampled from the expert. Subsequently, at iteration $i$, we collect data $D^i = \{(s, \pi^*(s)) | s \sim \pi^{i-1}\}$, where the states are sampled from the previously-trained learner with policy $\pi^{i-1}$. By merging current and previous data, $D^i \bigcup D^{i-1}$ is used to train new policy $\pi^i$. Intuitively, at each iteration, we collect a set of states that the learner is likely to visit based on previous experience, and obtain modified actions from the expert.

\begin{algorithm}[t]
\caption{Hierarchical Learn-form-Intervention DAgger} 
\hspace*{0.02in} {\bf Input:} parameters $L$, $T$, and policies $\mu$, $\pi_g , \forall g \in \mathcal{G}$ \\
\hspace*{0.02in} {\bf Output:} updated policies $\mu$, $\pi_g , \forall g \in \mathcal{G}$
\begin{algorithmic}[1]
\State Initialize data buffers $D_{h} \gets \emptyset $ and  $D_{g} \gets \emptyset $, $\forall g \in \mathcal{G}$ 
\Repeat
  \State Get a new environment instance with start state $s_0$
  \State Initialize data queue $\delta \gets \emptyset$
  \For{$ t= 1,\dots,T$}
    \State Get an updated state $s_t$
    \If{$\mu(s_t) = \mu^*(s_t)$}
      \If {no Intervention}
        \State Execute $a=\pi_g(s_t)$
        \If{ $len(\delta) > L$} POP$(\delta)$ \EndIf 
        \State Append $(s_t,a)$ to $\delta$
      \Else{}
        \State Execute $a=\pi^*(s_t)$
        \State $\delta \gets Backtrack(\delta, a)$
        \State Append $D_g \gets D_g \bigcup \{\delta\}$
        \State \textbf{Go to line 3}
    　\EndIf
　　\Else{}
　　　\State Append  $D_h \gets  D_h \bigcup \{(s_t,\mu^*(s_t)\}$
　　\EndIf
  \EndFor
\Until End of collecting 
\State Update meta-controller $\mu \gets $Train$(\mu,D_h)$ 
\State Update sub-policies $\pi_g \gets $Train$(\pi_g,D_{g})$ for all $g$
\end{algorithmic}
\label{alg:hlid}
\end{algorithm}

\subsubsection{Learn-from-Intervention} 
DAgger works fine for learning policies to play video games; however, adapting it to real world is difficult since we cannot allow robot to execute actions generated by an imperfect policy, which might jeopardize the safety of pedestrians on the road. It is also impractical to run the robot in pedestrian-rich environment before the model finally achieves good performance.

In real-world setting, we need to make frequent interventions to the robot and take control over the robot from early steps leading to a mistake. Such requirements make it hard to collect meaningful data trajectories $D^i = \{(s, \pi^*(s)) | s \sim \pi^{i-1}\}$ as in the original DAgger. To overcome the difficulties, we propose Learn-from-Intervention DAgger (see Algorithm~\ref{alg:hlid}), which allows the robot to learn how to react when it is likely going to make a mistake and also allows frequent expert interventions. 

The basic idea is inspired by the original DAgger, we let the learner controlled by itself and generate $s \sim \pi^{i-1}$, and the expert intervenes when the robot is likely going to make mistakes, e.g., get off the road or cause damage.
The expert makes an intervention and the control switches to the expert. For example, when the robot is going to drive off the road, the expert will make a steer and bring it back to the correct track. After making the correction, the control switches back to the robot. In contrast to the original DAgger, we record data a certain period $L$ before the expert intervention with a fixed length queue $\delta$. The reason why we need to record a period before the intervention are two fold. First, even the expert needs reaction time for the intervention. Second, the mistake taking place now is likely caused by more than one step rolling back the time. Therefore, the newly added data $\delta$ consists of the trajectory that is likely leading to a mistake $\{(s_p, \pi^{(i-1)}(s_p)) | p=t-1,\dots,t-L\}$ and the intervention of the expert $(s_t, \pi^*(s_t))$. 

In order to use the data in $\delta$, a backtrack function $Backtrack(\delta, a)$ is used to modify the state-action pairs in the buffer, which will later be added to the dataset $D_g$ after each intervention. At intervention time step $t$, the error is measured by the difference between the action $\pi_g (s_t)$ and $\pi^* (s_t)$. We test three different functions for backtracking the errors for $\{(s_p, \pi^{(i-1)}(s_p)) | p=t-1,\dots,t-L\}$ including logarithm, exponential and linear functions. Empirically, we find the linear function works best in our experiment. Furthermore, by learning from interventions after the robot gets warm start, the robot is easier to generalize what is right and what is wrong, therefore it gets better performance for the next iteration. Finally, we expend this framework to hierarchical structure.


\begin{figure}[t]
  \centering
  \includegraphics[width=\linewidth]{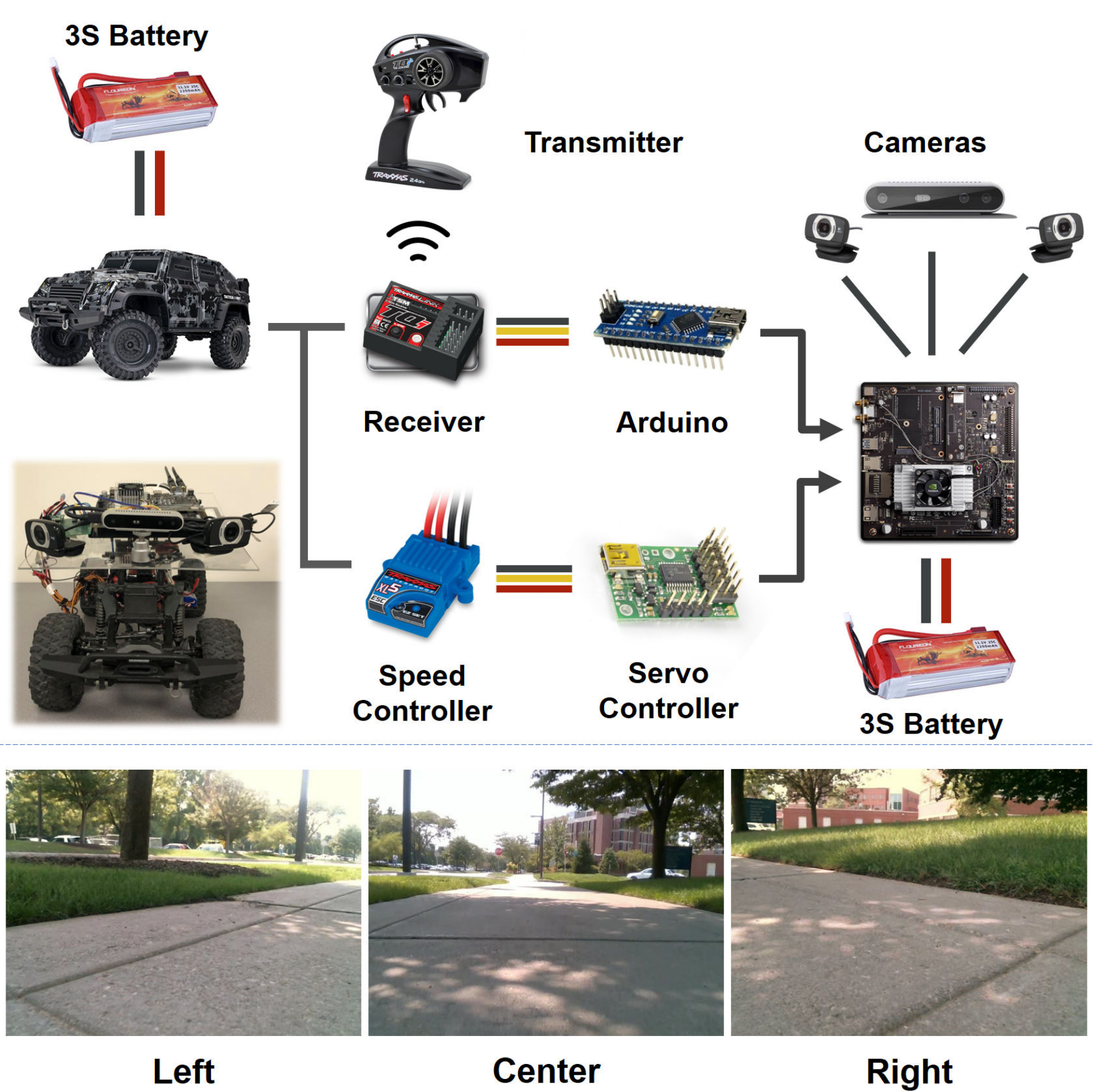}
  \caption{General graph of our hardware system setup indicating how wires connect and frames captured by three cameras in one spot. This three-camera setting provide a wide field of view to ensure the sides of paths can be recorded.}
  \label{fig:hardware}
\end{figure}


\section{SYSTEM}
\label{sec:system}

\subsection{Hardware Setup}
We evaluate our approach using an off-the-shelf RC car (Traxxas Tactical) as our robot platform, an NVIDIA Jetson TX2 as our main console, an Intel RealSense D415 as the main central camera and two webcams (LogiTech C615) on the sides. 
We also employ an Arduino board and a USB servo controller to implement our control system to the robot. Fig.~\ref{fig:hardware} shows our hardware setup. 

When performing imitation learning, the crucial part is how to collect the training data. In our case, the simplest solution is to collect every states when expert driving the robot in real world. However, this typically leads to imperfect policy, since the robot is only trained on the states encountered by the expert and will not recovery from mistakes when encountering an unseen state. To alleviate this problem, we include observations of recoveries from perturbations. Inspired by the pioneering work of Pomerleau~\cite{pomerleau1989alvinn}, we install three cameras in front of the car: one facing forward and the other two shifted to the left and to the right. By recording from three cameras simultaneously with appropriately adjusted control signals, we can simulate recovery from drift.

\subsection{Framework Overview}
\begin{figure}[t]
  \centering
  \includegraphics[width=\linewidth]{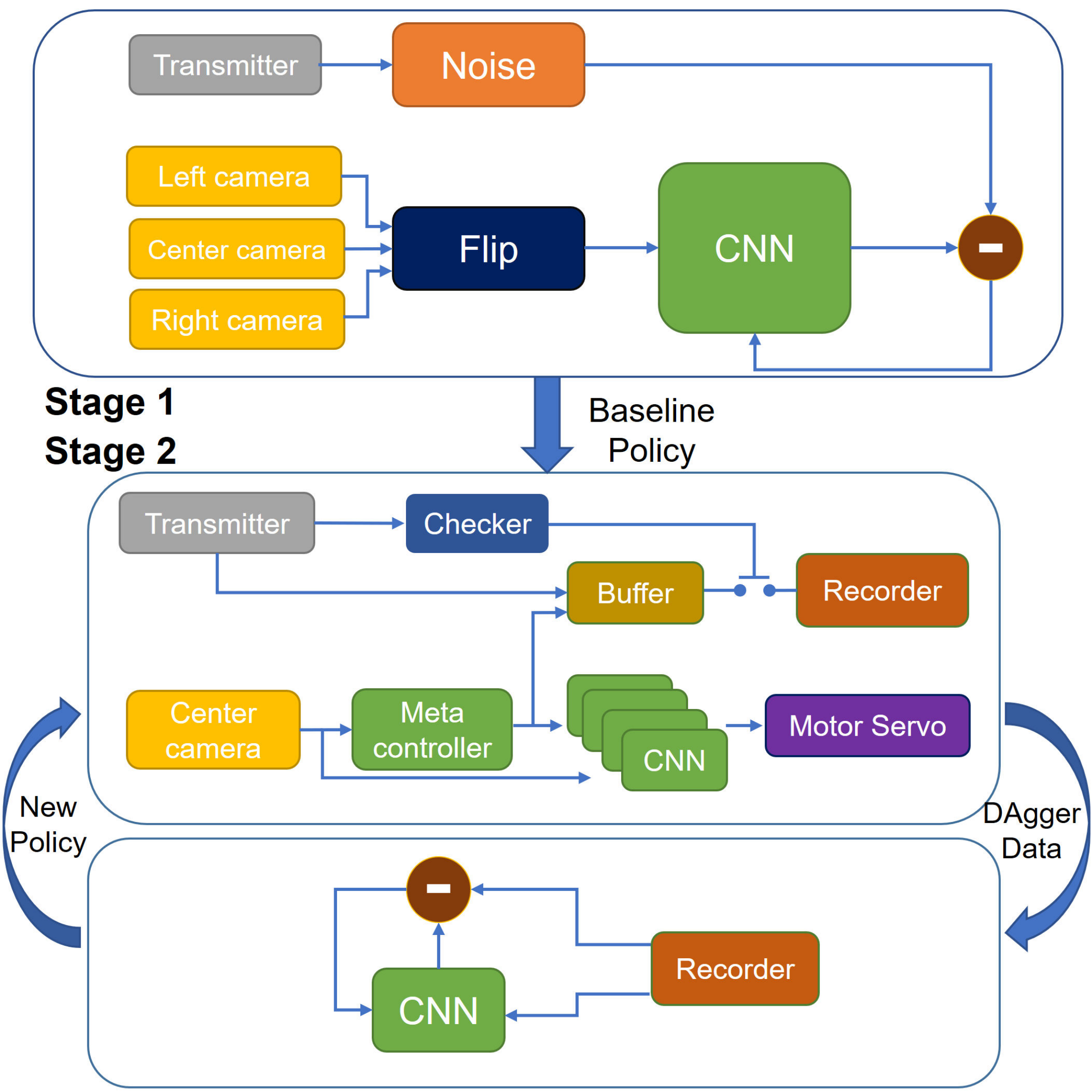}
  \caption{Framework overview, employing a hierarchical structure based on two separate stages. Two modules in Stage 2 interact iteratively to update our policy. Minus signs indicate the loss calculated by the loss function.}
  \label{pic:overview}
\end{figure}

The framework is depicted in Fig.~\ref{pic:overview}. Here a hierarchical framework containing two different stages is employed. We start stage 1 as warm-up stage which is used to generate a baseline policy for input to stage 2.
Given the baseline policy provided by warm-up stage, two separate modules work interactively in Stage 2. The learner takes policies as input, successively generates and executes actions based on the observations captured by the center camera. Once a human expert makes an intervention to the predicted trajectory, due to a violation of the policy based on the experts judgment, all data queued in buffer will be saved to $D^i$. We merge all data together to train a new policy $\pi^i$, which will be feedback to the learner in the next loop.

\subsubsection{policy representation}
For policy representation, we use a 2D CNN to map states to actions without considering sequential information for simplicity. We use transfer learning to adapt a pre-trained ResNet~\cite{He_2016_CVPR} on ImageNet to our data. We retrain the last few layers and retain the weights for other layers, because we think that the lower layers contain kernels extracting basic image features, e.g., lines and corners, and they are general for all images. Notice that we impose a basic assumption here: for similar observations, the actions of the expert should obey a single-peak distribution. In this case, we use cross-entropy as our loss function, which calculates the distance between the predicted distribution of the network and the actual action distribution of the expert. In this way, the network can learn more effectively of the expert actions than a one-hot representation that is used for other problems, e.g., image classification. 

\subsubsection{Stage 1}
We prepare our dataset by driving the RC car around the campus and recording featured environments (bituminous street, wide/narrow footpath, etc.). All movements of the car at this stage are controlled by a human expert, recorded in terms of PWM-signal-based tuples, and saved together with the video data into a SD card on Jetson TX2. We limit PWM command for throttle controlling from transmitter to below 1750, mapping to 1.2 ~ 1.5m/s in real world, in order to imitate the walking speed of human pedestrians and also to handle unexpected scenarios during testing. We also linearly map the PWM signals for steering control to actual steering angles, ranging from -50 to 50 degrees. Under our system, all saved commands are passed through TX2 for convenient data manipulation and calculation. Furthermore, we use horizontal flips to further augment the training dataset, and also flip the command based on the center steering angel with normal distribution noise. This provides the controller with demonstrations of recovery from drift and unexpected disturbances.

\subsubsection{Stage 2} 
We use all data collected from Stage 1 to get a baseline policy. However, we have found that the policy learned with Stage 1 are not sufficiently robust. The robot still make mistakes at the training region where we use for collecting the training data. It performs even worse in unseen testing region, because only using states encountered by the expert is not enough for a good generalization. Therefore, we further use our Learn-from-Intervention DAgger algorithm to provide more crucial data to train our policies. To do so, we set up our platform to be able to record and inference at the same time. We implement a fixed length queue to retain the seen states and predicted commands. If there is no invention from the expert, the queue will pop out unused data. Upon detecting a likely mistake, the expert will correct the robot and the queue will be used. The model is evaluated on the TX2 in real time (inference performed at 10 FPS). It receives images from the central camera. The network predicts steer angle and speed value control command, which will be converted into appropriate PWM signals. After conversion, signals will be directly seed to the speed controller and steering servo.




\section{EXPERIMENTS}
\label{sec:exp}

\begin{figure}[t]
\centering
	\includegraphics[width=\linewidth]{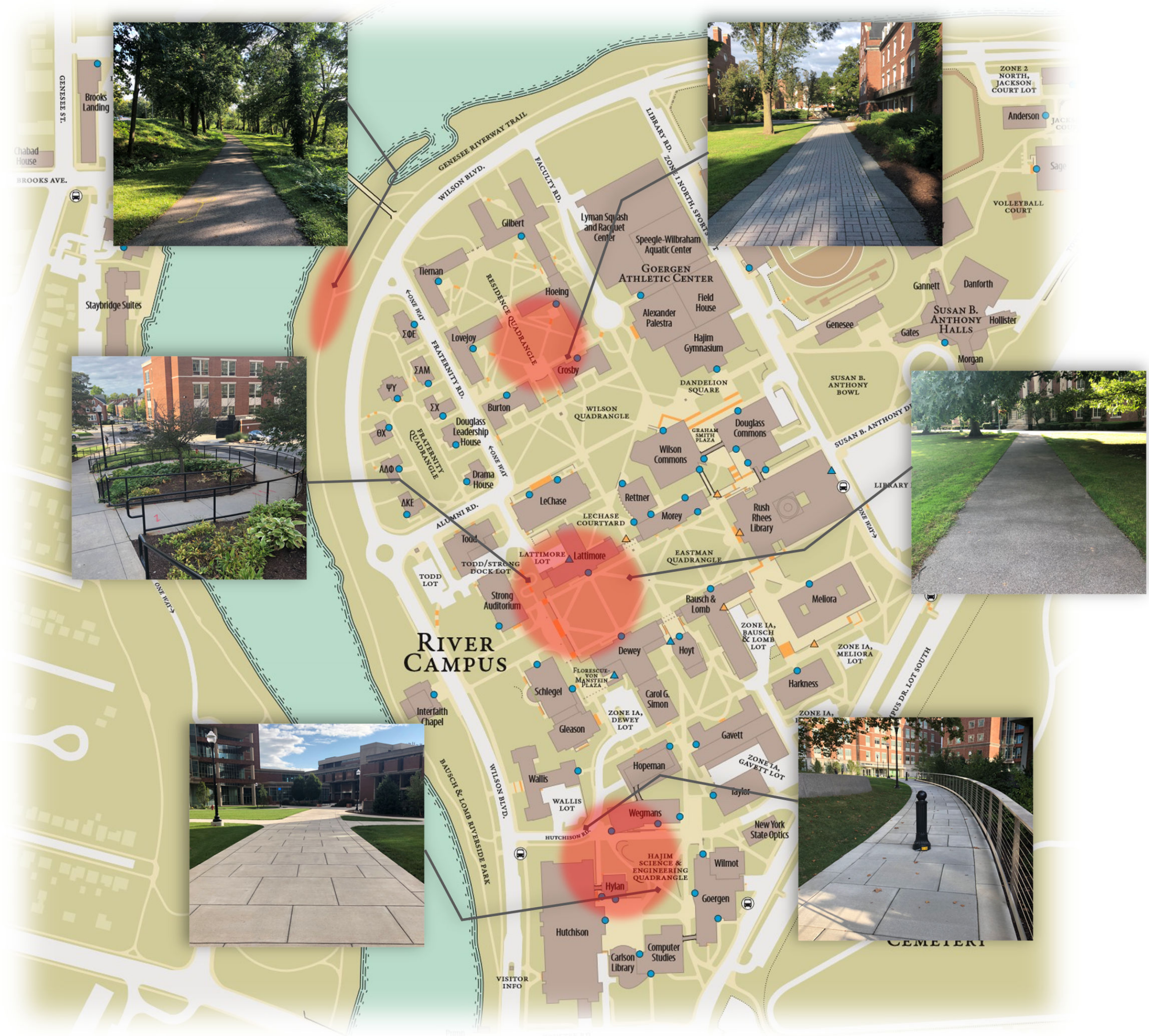}
	\caption{A map of our campus, with red areas indicating our training and testing areas. Pictures around the map present some featured examples of the real-world views of these areas.}
	\label{pic:campus}
\end{figure}

To eliminate the influence of training data and evaluate qualitatively how well the model generalizes to previously unseen environments with different appearance, we intentionally choose four regions excluded by our dataset as our experimental field which are marked as red in Fig.\ref{pic:campus}. 
For each sub-task, we use behavior cloning model as our first iteration and generate five policies which differ in the number. 

During testing process, for each policy in every region, the performance of the robot is characterized as the proximity of its behaviors to human expert's actions using two types of measurements for different sub-tasks.

\label{sec:exp}

\subsection{Successful Attempts}

\begin{figure}[t]
\centering
	\includegraphics[width=\linewidth]{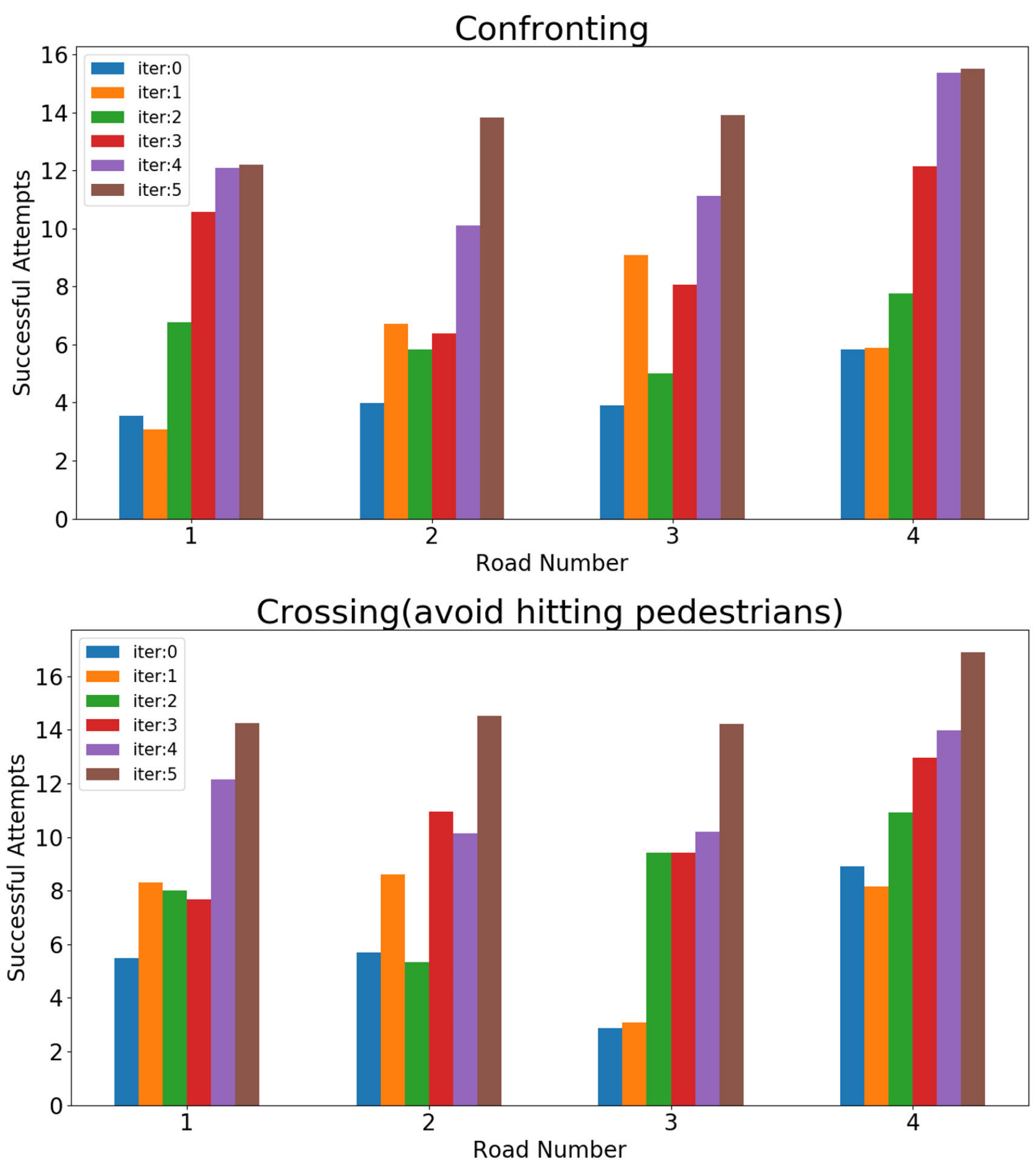}
	\caption{Successful attempts of six models for pedestrian following and path following on each of for testing road within every 20 attempts. X-axis indicates the number of each road and the Y-axis indicates numbers of the successful trials. }
	\label{pic:class2_3}
\end{figure}

One of the most intuitive way to assess the achievements of our robot is to measure the number of successful attempts. For scenarios such as confronting and crossing, we just adopt this straightforward approach as our evaluation method. For individual sub-task, we make in total 20 attempts on four different roads for each model we trained. If the robot successfully shun confronting pedestrians or stop for crossing pedestrians, we will count that as a success. In Fig. \ref{pic:class2_3}, it presents the number of successful attempts for each policy on each road.
As shown in Fig.\ref{pic:class2_3}, robot achieve more successful attempts after each iteration of learning from intervention which indicates the improvement on generalization of the policy.

\subsection{Time Without Intervention}
\begin{figure}[t]
\centering
	\includegraphics[width=\linewidth]{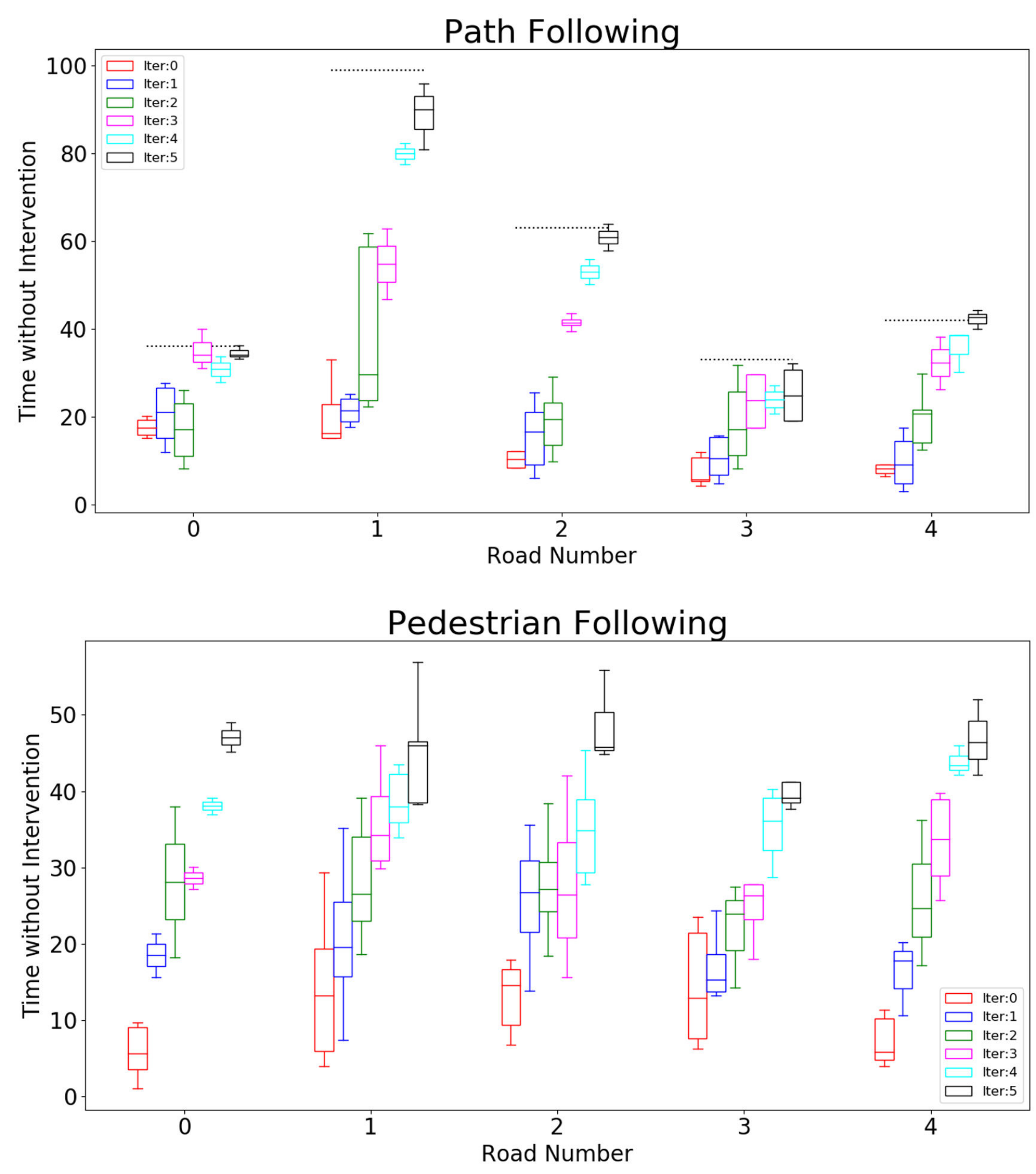}
	\caption{TWI of six models for confronting and crossing on five different roads. We show the results in terms of box plots.}
	\label{pic:class0_1}
\end{figure}

Apart from those two tasks mentioned in the previous section, we also evaluate the competence of our robot on two other sub-tasks, path following and pedestrian following, in which the performance could not be precisely quantified in terms of number of successful attempts since the task of following is an enduring process rather than one single action. Therefore, we choose to measure the time without intervention (TWI) which means the duration of the robot successfully following the object in autonomous mode. Results for experiments on four different roads are shown in Fig. \ref{pic:class0_1}. For each road, there is a gray line representing the total time our human needs to walk from one end to the other. Every box denotes the TWI of all five experiments.
The result indicates that generalization of our model is indeed enhanced after iteratively learning from human intervention.
\subsection{Training Data}
\begin{figure}[t]
\centering
	\includegraphics[width=\linewidth]{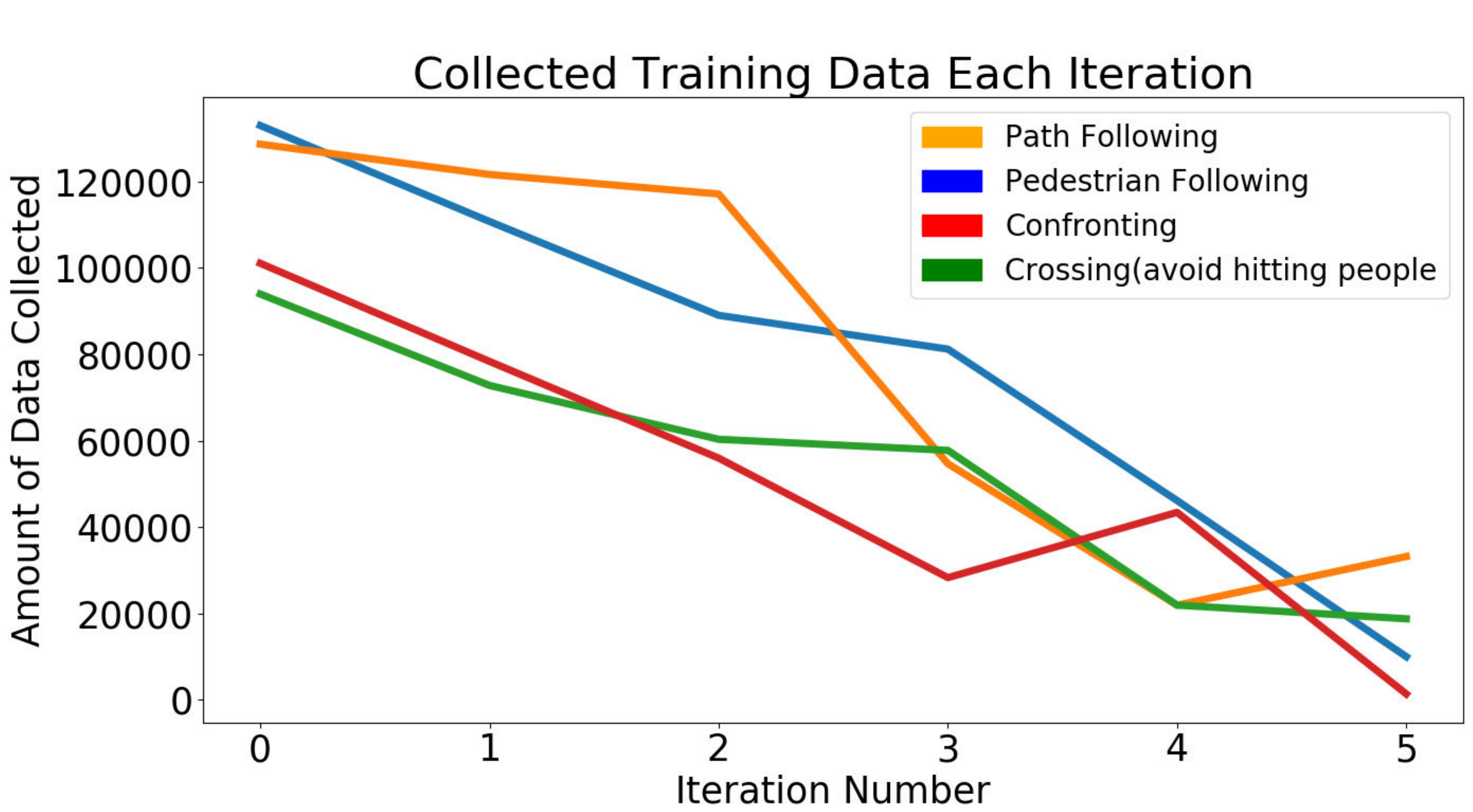}
	\caption{Changes in size of dataset related to number of iterations for all four scenarios. For each loop, the increments of data collected during DAgger process fall due to less errors made by the robot under updated policies.}
	\label{pic:data}
\end{figure}
Since our framework merely take failures into consideration, as our model improves, the amount of data we could collect dramatically decreases in a pattern exhibited in Fig. \ref{pic:data}. We found that after five iterations the size of our dataset for each subtask will converge to a saturated state and basically remains invariant.

\section{CONCLUSIONS}
\label{sec:conclusions}

To conclude, we introduce the learn-from-intervention DAgger algorithm, an approach for robot to navigate in a pedestrian-rich environment from raw images. We apply the proposed approach to a physical robotic vehicle in real-world road with pedestrians. The result of experiment indicates that imitating human behavior from intervention is useful for robot to generalize a good decision.
For the future work, we plan to model this problem as a sequence-decision problem and take previous actions and observations into consideration when training a policy representation, which makes it more realistic for robots to generate actions.

\bibliographystyle{IEEEtran}

\end{document}